\documentclass[11pt]{article}

\usepackage{acl}
\usepackage{times}
\usepackage{latexsym}

\usepackage[T1]{fontenc}
\usepackage[utf8]{inputenc}
\usepackage{microtype}
\usepackage{inconsolata}

\usepackage{graphicx}
\usepackage{amsmath} 

\usepackage{multirow}  
\usepackage{graphicx}  
\usepackage{booktabs}  
\usepackage{makecell}

\newcommand{\head}[2]{\makecell[c]{\textit{#1} \\ \textit{#2}}}

\newcommand{\Ins}{Instruct }
\newcommand{\Thi}{Thinking }

\newcommand*{\affmark}[1][*]{\textsuperscript{#1}}
\usepackage{fdsymbol}

\title{Revisiting Model Interpolation for Efficient Reasoning}

\author{
Taiqiang Wu\affmark[$\diamondsuit$] \ Runming Yang\affmark[$\diamondsuit$] \ Tao Liu\affmark[$\spadesuit$] \ 
Jiahao Wang\affmark[$\diamondsuit$] \ Ngai Wong\affmark[$\diamondsuit$]
\\
\affmark[$\diamondsuit$]The University of Hong Kong \ 
\affmark[$\spadesuit$]Tsinghua University
\\
{\tt takiwu@connect.hku.hk} \ {\tt nwong@eee.hku.hk}
}

\begin{document}
\maketitle
\begin{abstract}
Model merging, typically on \Ins and \Thi models, has shown remarkable performance for efficient reasoning.
In this paper, we systematically revisit the simplest merging method that interpolates two weights directly.
Particularly, we observe that model interpolation follows a three-stage evolutionary paradigm with distinct behaviors on the reasoning trajectory.
These dynamics provide a principled guide for navigating the performance-cost trade-off. 
Empirical results demonstrate that a strategically interpolated model surprisingly surpasses sophisticated model merging baselines on both efficiency and effectiveness.
We further validate our findings with extensive ablation studies on model layers, modules, and decoding strategies.
Ultimately, this work demystifies model interpolation and offers a practical framework for crafting models with precisely targeted reasoning capabilities.
\end{abstract}

\section{Introduction}

Large language models~(LLMs), such as Qwen3~\citep{yang2025qwen3}, OpenAI o1~\citep{jaech2024openai}, and Deepseek R1~\citep{guo2025deepseek}, have revolutionized the field of natural language processing~(NLP).
Their remarkable success in complex tasks is largely attributed to emergent reasoning capabilities, which benefit from scaled chain-of-thoughts~\citep{wei2022chain} during test time~\citep{snell2024scaling}.
However, longer CoT also introduces significant trade-offs such as over-thinking~\citep{chen2024not} and high latency issues~\citep{sui2025stop}.
Consequently, how to achieve efficient reasoning without compromising performance remains a critical challenge.

To address this challenge, model merging has emerged as a compelling solution~\citep{yang2024model,wu2025timber}
The core idea is to merge the weights of two specialized models, including a \Thi model adept at long-CoT reasoning and an \Ins model optimized for short, direct answers, and thus create a hybrid model balancing reasoning capability with token efficiency~\citep{team2025kimi, wu2025unlocking}.
Existing merging methods can be categorized into weighted-based~\citep{utans1996weight, ilharco2022editing}, subspace-based~\citep{yadav2023resolving,yu2024language}, and routing-based~\citep{muqeeth2023soft,tang2024towards}.
Most of these merging methods require the paired pretrained models, a constraint not shared by the simpler method of direct model interpolation.

\begin{figure}[t]
  \includegraphics[width=\linewidth]{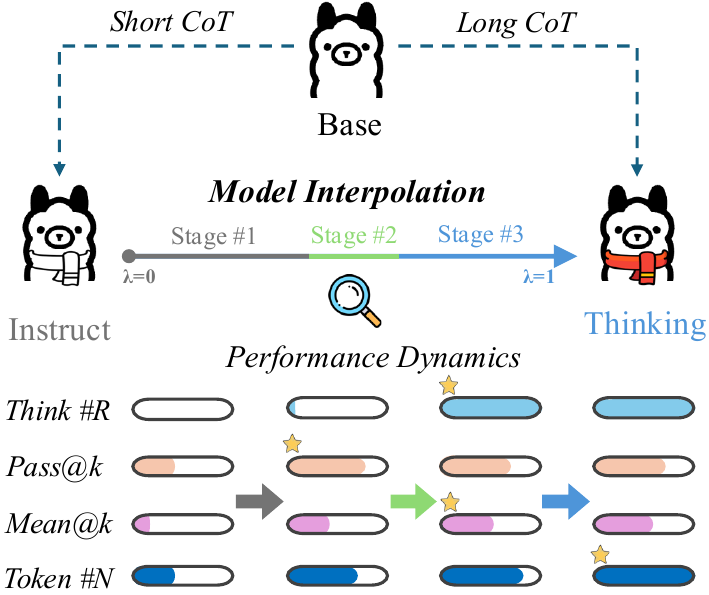}
  \caption{The performance dynamics for the model interpolation between \Ins and \Thi models.
  \textit{Think \#R} denotes the ratio of samples with \textit{</think>} token in responses.
  \textit{Token \#N} denotes the number of tokens in responses.}
  \label{fig:framework}
\end{figure}

\begin{figure*}[t]
\centering
  \includegraphics[width=0.95\linewidth]{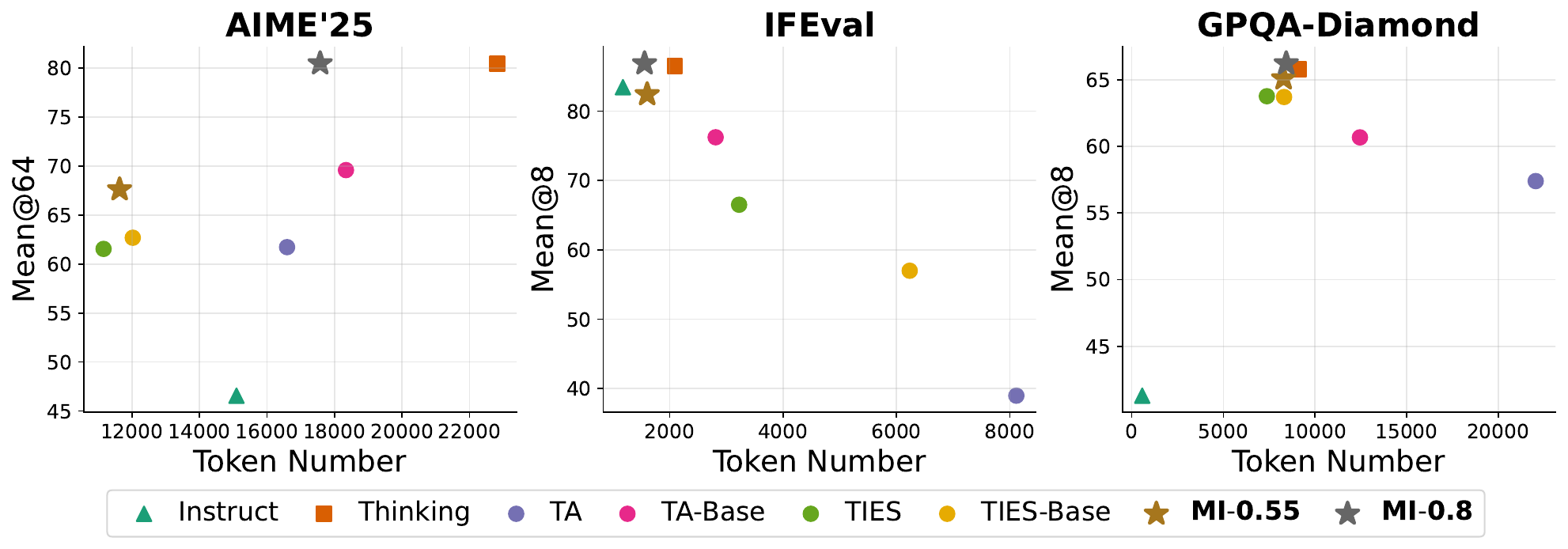}
  \caption{Performance of vanilla Instruct, Thinking, and model merging methods on AIME'25, IFEval, and GPQA-Diamond.
  \textbf{MI} denotes the model interpolation and the suffix for the interpolation coefficient $\lambda$.
  The results indicate that \textbf{MI} surpasses these baselines on both efficiency and effectiveness.}
  \label{fig: 4B_scatter}
\end{figure*}

In this paper, we revisit the model interpolation~(\textbf{MI}) method and systematically analyze the dynamics as the interpolation coefficient $\lambda$ sweeps from 0 to 1.
Surprisingly, we find that the performance metrics, such as Pass@k and Mean@k, do not evolve linearly but instead follow a distinct three-stage paradigm detailed in Figure \ref{fig:framework}.
In stage \#1, where the weights of the Instruct model are dominant, Pass@k and the number of output tokens increase rapidly with \textbf{almost no} explicit reasoning in responses.
The thinking pattern, measured by the presence of token \textit{</think>}, emerges in stage \#2 with larger $\lambda$, with the Mean@k increasing faster than Pass@k.
At stage \#3, the output responses become substantially longer, yet the marginal gains in Pass@k and Mean@k diminish, corresponding to the overthinking phenomenon~\citep{chen2024not}.
These observations offer novel insights into the mechanics of weight interpolation and further provide a principled guide for desired reasoning behaviors.

We conduct extensive experiments on Qwen3 models~\citep{yang2025qwen3}, interpolating between the official \Thi and \Ins variants.
As shown in Figure \ref{fig: 4B_scatter}, a strategically interpolated model surpasses the mainstream model merging baselines across various challenging benchmarks, including mathematical reasoning~(AIME'25~\citep{AIME2025}), instruction-following~(IFEval~\citep{zhou2023instruction}), and science reasoning~(GPQA-Diamond~\citep{rein2024gpqa}).
Beyond this superior performance, our analysis provides a practical framework for crafting models with desired behaviors such as adhering to a specific token budget.
We further conduct fine-grained ablations on model layers and modules, providing a comprehensive view of how interpolation fuses reasoning capabilities.

Our main contributions can be summarized as follows:
\begin{itemize}
    \item We systematically revisit model interpolation methods and uncover a three-stage evolutionary paradigm. 
    This framework provides principled guidance for efficient reasoning.
    \item We empirically demonstrate that a simple yet strategically interpolated model can surpass sophisticated merging baselines across a diverse suite of challenging benchmarks. 
    \item We provide deep mechanistic insights through fine-grained ablation studies on layers, modules, and decoding strategies.
\end{itemize}

\section{Related Work}
\subsection{Efficient Reasoning}

Efficient reasoning aims to reduce the output tokens while preserving reasoning capability as much as possible~\citep{sui2025stop}.
The methods can be categorized as 1) model-based, setting a short response as the optimization goal during SFT~\citep{yu2025z1, li2025tl} or RL~\citep{team2025kimi, zhang2025adaptthink, zhang2025making}, 2) decoding-based, which modifies the output paradigm such as latent reasoning~\citep{hao2024training,butt2025soft} and dynamic reasoning~\citep{sun2024fast, wang2025sampling}, and 3) prompts-based, refining the input prompts for enable concise and efficient
reasoning~\citep{xu2025chain,aytes2025sketch}.
We refer the reader to \citet{sui2025stop} for a comprehensive survey.
In this paper, we focus on the training-free model interpolation for efficient thinking.

\subsection{Model Merging}

Model merging methods merge the parameters of multiple separate models with different capabilities, and are widely applied for continual learning~\citep{marczak2024magmax}, multi-task learning~\citep{yang2023adamerging}, and model attack~\citep{gangwal2025merge}.
The key is to merge the weights from different models following the same structure.
One intuitive idea is to average the weights~\citep{utans1996weight}, while the task arithmetic framework extends the idea to the task vector~\citep{ilharco2022editing}.
Please refer to \citet{yang2024model} for more details.
Recently, Kimi k1.5 averages two models with long CoT and short CoT capabilities for efficient reasoning~\citep{team2025kimi}, which can be viewed as a special case of model interpolation with coefficients $\lambda$ being 0.5. 
In this work, we extend such idea and analyze the model interpolation with various coefficients. 
\section{Preliminary}

\subsection{Instruct and Thinking Models}

Regarding the reasoning ability of LLMs, there are two distinct reasoning styles, i.e., long CoT and short CoT.
Typically, long CoT~(a.k.a, slow thinking) LLMs perform better on complex tasks with an explicit thinking process but incur greater inference latency, while short CoT~(a.k.a, quick thinking) LLMs are optimized to produce short, direct answers, prioritizing speed and efficiency.

These two reasoning modes are typically realized through two primary paradigms.
The first one is a hybrid reasoning supporting both Thinking and Non-thinking,  such as \texttt{Qwen3-4B}~\citep{yang2025qwen3}.
The second paradigm involves post-training two separate specialist models, such as \texttt{Qwen3-4B-Instruct-2507} for quick thinking and \texttt{Qwen3-4B-Thinking-2507} for slow thinking.

To better understand the relationship between these model variants, we analyze their parameter similarity by the $\sigma$ defined in~\citet{wu2025shadow}.
Table~\ref{table:4B-sigma} reports the results among Qwen3-4B series, revealing two key findings.
First, all the paired models are highly similar in weights.
Second, \texttt{Qwen3-4B} is more similar to \texttt{Qwen3-4B-Thinking-2507} models than \texttt{Qwen3-4B-Base}, suggesting a potential inheritance relationship.

\begin{table}[!t]
\centering
\resizebox{\linewidth}{!}{
\begin{tabular}{l|l|c}
\toprule
  \multicolumn{2}{c|}{{\bf Models}} & $\sigma$ \\
\midrule
    \texttt{Qwen3-4B-Base} & \texttt{Qwen3-4B} & 0.0326 \\
    \texttt{Qwen3-4B-Base}& \texttt{Qwen3-4B-Instruct} & 0.0562 \\
    \texttt{Qwen3-4B-Base}& \texttt{Qwen3-4B-Thinking} & 0.0638 \\
    \texttt{Qwen3-4B}& \texttt{Qwen3-4B-Instruct} & 0.0589 \\
    \texttt{Qwen3-4B}& \texttt{Qwen3-4B-Thinking} & 0.0633 \\
    \texttt{Qwen3-4B-Instruct}& \texttt{Qwen3-4B-Thinking} & 0.0269 \\
\bottomrule
\end{tabular}
}
\caption{Weight similarity $\sigma$ \citep{wu2025shadow} on paired models from Qwen3 series.
We omit the suffix \texttt{-2507} for simplicity.
The smaller $\sigma$, the more similar.}
\label{table:4B-sigma}
\end{table}

\subsection{Model Merging Methods}
\label{sec:merge_baseline}

Given a set of $T$ models, $\{\Theta^{(1)}, \dots, \Theta^{(T)}\}$, that share a common architecture and are typically trained based on the same pre-trained model $\Theta^{(0)}$, the objective of model merging is to produce a new model, $\Theta^{(\text{Merge})}$, through a parameter-wise operation:
\begin{equation}
    \Theta^{(\text{Merge})} = \text{Merge}(\Theta^{(0)},\Theta^{(1)}, \dots, \Theta^{(T)}).
\end{equation}
An intuitive baseline strategy is to average the parameters of the models:
\begin{equation}
    \Theta^{(\text{Merge})} = \frac{1}{T} \sum_{i=1}^{T} \Theta^{(i)}.
\end{equation}

\paragraph{Task Arithmetic.} \citet{ilharco2022editing} defines task vector~(TV) to represent the parameter shift on a specific task, which is calculated as:
\begin{equation}
    TV^{(i)} := \Theta^{(i)} - \Theta^{(0)}.
\end{equation}
Thus, multiple capabilities can be combined by aggregating their corresponding task vectors and adding them back to the base model:
\begin{equation}
\label{eq:tv}
    \Theta^{(\text{Merge})} = \Theta^{(0)} + \alpha \sum_{i=1}^{T} TV^{(i)},
\end{equation}
where $\alpha$ is the scaling factor for all TVs.

\paragraph{TIES-Merging.} TIES-Merging~\citep{yadav2023resolving} proposes to transform weights into sparse subspaces for merging due to the over-parameterized nature of neural networks~\citep{choudhary2020comprehensive}.
For a given task vector $TV^{(i)}$, TIES-Merging retains part of parameters with the highest magnitudes:
\begin{equation}
    TV^{(i)}_{\text{TIES}} = \text{Top-k}(TV^{(i)}).
\end{equation}
These sparse task vectors are then resolved by voting for signs and merged following Equation \ref{eq:tv}.

\paragraph{Model Interpolation (MI).} 
In this paper, we focus on the task of efficient reasoning by merging a Thinking model~$\Theta^{(\text{Thi})}$, and an Instruct model~$\Theta^{(\text{Ins})}$. 
The model interpolation (MI) is formulated as:
\begin{equation}
    \Theta^{(\text{Merge})} = \lambda \Theta^{(\text{Thi})} + (1-\lambda) \Theta^{(\text{Ins})}.
\end{equation}
This process is agnostic to the base model and can be framed as a special case of Task Arithmetic. 
By defining $TV^{(\text{Thi})} = \Theta^{(\text{Thi})} - \Theta^{(\text{Base})}$ and $TV^{(\text{Ins})} = \Theta^{(\text{Ins})} - \Theta^{(\text{Base})}$, it is easy to show that:
\begin{equation}
\begin{aligned}
    &\lambda \Theta^{\text{(Thi)}} + (1-\lambda) \Theta^{\text{(Ins)}}
    = \lambda (TV^{\text{(Thi)}} + \Theta^{\text{(Base)}}) \\ &+ (1-\lambda) (TV^{\text{(Ins)}} + \Theta^{\text{(Base)}}) \\
    &= \Theta^{\text{(Base)}} + \lambda TV^{\text{(Thi)}} + (1-\lambda) TV^{\text{(Ins)}}.
\end{aligned}
\end{equation}
The $\Theta^{\text{(Base)}}$ can be an \textit{arbitrary} model.
This derivation shows that MI is equivalent to performing Task Arithmetic on the \Thi and \Ins task vectors with scaling factors of $\lambda$ and $(1-\lambda)$, respectively.

\section{Revisiting Model Interpolation}

\subsection{Experimental Setup}

\paragraph{Models.}
We conduct experiments merging the Qwen3-4B and Qwen3-30B-A3B models.
For both settings, we download the official weights from huggingface and merge the \texttt{Instruct-2507} and \texttt{Thinking-2507} variants.

\paragraph{Benchmarks.}
To ensure a comprehensive evaluation, we select three representative benchmarks that cover diverse reasoning skills, containing IFEval for instruction following~\citep{zhou2023instruction}, GPQA-Diamond for scientific reasoning~\citep{rein2024gpqa}, and AIME'25 for mathematical reasoning~\citep{AIME2025}. 
We adapt the OpenCompass~\citep{2023opencompass} framework for evaluation.
Further details of these benchmarks are provided in Appendix~\ref{appendix: benchmark}.

\paragraph{Decoding Strategy.}
For the baseline \Ins and \Thi models, we employ their official sampling configurations to ensure optimal performance. 
The \Thi model uses a temperature $T$ of 0.6 and Top-p of 0.95, while the \Ins model uses 0.7 and 0.8, respectively. 
For all merged models, we consistently apply the same hyperparameters with \Thi model~(i.e., $T=0.6$, $\text{Top-p}=0.95$), deferring a detailed analysis of hyperparameter sensitivity to Section~\ref{sec:hyper}. 
We roll out $64$ times for AIME'25 and $8$ for IFEval and GPQA-Diamond.

\paragraph{Evaluation Metrics.}
We evaluate the models across the following abilities:
\begin{itemize}
    \item \textbf{Effectiveness.} 
    We report Pass@k and Mean@k scores. 
    For Pass@k, we adapt the unbiased estimator~\citep{chen2021evaluating}.
    \item \textbf{Consistency.}
    We also report Vote@k to measure the stability of the model's most frequent answer.
    
    \item \textbf{Efficiency.}
    We measure computational cost by the average number of tokens in the generated responses, denoted as Token \#N.
    
    \item \textbf{Reasoning Pattern.} 
    We further introduce the Thinking Ratio~(Think \#R), defined as the percentage of responses containing the \textit{</think>} token, to quantify the prevalence of explicit CoT reasoning.
\end{itemize}

\begin{figure*}[!t]
\centering
\includegraphics[width=\linewidth]{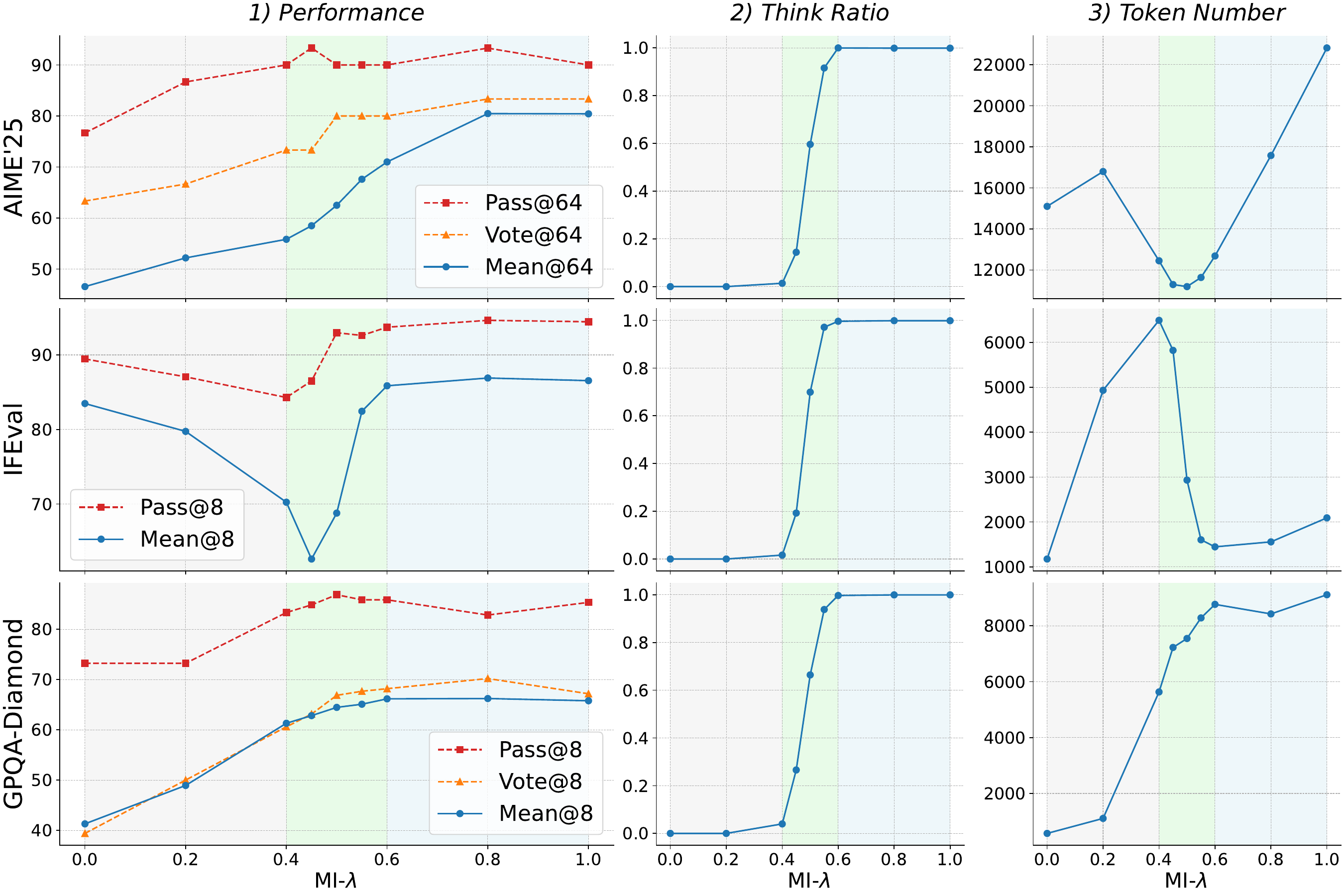}
  \caption{The performance dynamics of model interpolation~(MI) on \texttt{Qwen3-4B-Instruct-2507} and \texttt{Qwen3-4B-Thinking-2507}.
  The dynamics follow a three-stage evolutionary paradigm colored in \colorbox[HTML]{f6f6f6}{grey}, \colorbox[HTML]{ecfbea}{green}, and \colorbox[HTML]{f0f7fa}{blue}.
  $\lambda$ denotes the interpolation coefficient ranging from 0 to 1.
  Please refer to Appendix \ref{appendx: qwen30B} and Appendix \ref{appendx: llama3-8B} for results on \texttt{Qwen3-30B-A3B} and \texttt{Llama-3.1-8B}, respectively.
  }
  \label{fig:dynamics_4b}
\end{figure*}

\subsection{Three-Stage Paradigm}
\label{sec:paradigm}

Figure~\ref{fig:dynamics_4b} and Figure~\ref{fig:dynamics_30b} illustrate the performance dynamics on the Qwen3-4B and Qwen3-30B-A3B models, respectively. 
Across both model scales, structures, and all three benchmarks, we observe that the performance dynamics do not evolve linearly with the interpolation coefficient $\lambda$.
Instead, they follow a consistent and predictable three-stage paradigm, which we detail below using the Qwen3-4B model as the primary example.

\paragraph{Stage \#1.}
Corresponding to $\lambda \in [0, 0.4)$ for Qwen3-4B model.
In this initial phase, the merged model is dominated by the \Ins model but begins to incorporate traits from the \Thi model, and thus \textbf{generating longer outputs without adopting an explicit thinking process}.

The performance of different abilities distinct each other.
The Think Ratio (\textit{Think \#R}) remains near zero. 
Hence, the model almost never generates explicit Chain-of-Thought steps, opting for direct answers.
Meanwhile, the number of tokens (\textit{Token \#N}) and Pass@k gradually increase as the model begins to generate more verbose responses.
For example, the \textit{Token \#N} on IFEval increases from 1174 to 6492.
However, due to the lack of an explicit reasoning process, the Mean@k and Vote@k increase much more gently on AIME'25 and GPQA-Diamond.
In addition, there is a \textit{large drop} on IFEval, since some input questions require being answered with token limits.

\paragraph{Stage \#2.}
Corresponding to $\lambda \in [0.4, 0.6]$ for Qwen3-4B.
In this stage, the \textbf{reasoning pattern following \Thi models rapidly emerges}, leading to largely increased Mean@k and gently increased Pass@k and \textit{Token \#N}.
This stage marks a critical and dramatic phase transition.

Specifically, the \textit{Think \#R} abruptly rises from nearly 0 to 1, indicating the rapid emergence of explicit long CoT capabilities from the \Thi model.
Across the three benchmarks, all the metrics show gains in this stage. 
In contrast to stage \#1, the Mean@k scores increase largely while the Pass@k scores more gently.
For instance, the Mean@64 score increases from 55.8 to 71.0 on AIME'25, while Pass@64 is unchanged.
The sharp increase in Mean@k suggests the dramatically improved quality of the reasoning process. 
Notably, the Pass@k score often reaches its peak at the end of this stage.
This stage corresponds to a significant decrease in token count, indicating a potential sweet spot for effectiveness and efficiency.

\paragraph{Stage \#3.}
Corresponding to $\lambda \in (0.6, 1.0]$ for Qwen3-4B.
In this final stage, the merged model \textbf{converges} to the pure \Thi model, with \textbf{continuously increasing} \textit{Token \#N} and slight change in Pass@k and Mean@k.

At this stage, the \textit{Think \#R} is saturated at 1.0 and the \textit{Token \#N} continuously increases, reflecting the high cost of generating long-form reasoning for all inputs.
Although Mean@k continues to show slight improvements, Pass@k often plateaus or even slightly declines from its peak at Stage \#2. 
This suggests a point of diminishing returns and provides clear evidence of the over-thinking phenomenon~\citep{chen2024not}, where longer reasoning does not necessarily lead to better performance.
Interestingly, the merged model can outperform the pure \Thi model ($\lambda=1.0$) at certain points in this stage (e.g., at $\lambda=0.8$), demonstrating that a slight blend with the \Ins model can sometimes regularize the reasoning process and yield better performance~\citep{wu2025timber}.

\paragraph{Discussion on Larger Model.}
The performance dynamics of the much larger Qwen3-30B-A3B models (shown in Figure~\ref{fig:dynamics_30b}) follow a similar three-stage paradigm, confirming the generalization of our findings. 
However, the specific ranges for each stage differ, with Stage \#2 occurring later, at $\lambda \in [0.5, 0.8]$.
Experiments on Llama-3.1-8B indicate a consistent conclusion across different stages.
Further details are available in Appendix~\ref{appendx: qwen30B} and Appendix~\ref{appendx: llama3-8B}.

\begin{table*}[!t]
\centering
\renewcommand{\arraystretch}{1.1} 
\resizebox{\textwidth}{!}{
\begin{tabular}{lccccccccc}
\toprule
\multirow{3}{*}{\textbf{Method}} & \multicolumn{3}{c}{\textbf{AIME'25}} & \multicolumn{3}{c}{\textbf{IFEval}} & \multicolumn{3}{c}{\textbf{GPQA-Diamond}} \\
\cmidrule(lr){2-4}
\cmidrule(lr){5-7}
\cmidrule(lr){8-10}
& \head{Mean/Pass}{@64} & \head{Token}{\#N} & \head{Think}{\#R} &
  \head{Mean/Pass}{@8} & \head{Token}{\#N} & \head{Think}{\#R} &
  \head{Mean/Vote}{@8} & \head{Token}{\#N} & \head{Think}{\#R} \\
\midrule
Instruct     & 46.6/76.7 & 15097 & 0.0   & 83.5/89.5 & 1174  & 0.0   & 41.3/39.4 & 570   & 0.0   \\
Thinking     & 80.4/90.0 & 22813 & 99.9  & 86.6/94.5 & 2091  & 100.0 & 65.8/67.2 & 9114  & 99.9  
\\
\midrule
TA           & 61.7/86.7 & 16594 & 0.5   & 39.0/60.8 & 8116  & 7.6   & 57.4/59.6 & 22042 & 0.2   \\
TA-Base      & 69.6/86.7 & 18339 & 43.0  & 76.3/91.9 & 2810  & 76.5  & 60.7/66.2 & 12450 & 32.9  \\
TIES         & 61.5/93.3 & 11159 & 65.4  & 66.5/90.8 & 3224  & 71.2  & 63.8/68.2 & 7369  & 72.2  \\
TIES-Base    & 62.7/90.0 & 12024 & 15.6  & 57.0/87.8 & 6234  & 21.6 &  63.7/66.7 & 8309  & 21.8 \\
DARE & 61.4/93.3 & 11231 & 38.0 & 62.8/91.7 & 2758 & 51.7 & 64.4/67.7 & 7215 & 58.4 \\
SLERP & 62.6/90.0 & 10944 & 60.4 & 68.7/92.1 & 3220 & 68.2 & 65.2/69.2 & 7523 & 65.2
\\
\midrule 
MI-0.2        & 52.2/86.7 & 16794 & 0.0   & 79.7/87.1 & 4933  & 0.0   & 48.9/50.0 & 1107  & 0.0   \\
MI-0.4        & 55.8/90.0 & 12448 & 1.4   & 70.2/84.3 & 6492  & 1.6   & 61.3/60.6 & 5634  & 4.0   \\
MI-0.5        & 62.5/90.0 & 11189 & 59.6  & 68.8/93.0 & 2932  & 69.9  & 64.5/66.7 & 7548  & 66.4  \\
MI-0.6        & 71.0/90.0 & 12681 & 100.0 & 85.9/93.7 & 1445  & 99.6  & 66.2/68.2 & 8769  & 99.7  \\
MI-0.8        & \textbf{80.5/93.3} & 17574 & 99.9  & \textbf{86.9/94.6} & 1556  & 99.9  & \textbf{66.2/70.2} & 8427  & 99.9  \\
\bottomrule
\end{tabular}
}
\caption{Performance comparison across AIME'25, IFEval, and GPQA-Diamond when merging \texttt{Qwen3-4B-Instruct-2507} and \texttt{Qwen3-4B-Thinking-2507}.
We sample 64 times on AIME'25 and 8 times on others.
Since questions from GPQA-Diamond are multiple-choice, we report Vote@8 instead of Pass@8.
}
\label{tab:main_res}
\end{table*}

\subsection{Compared with More Baselines}
\label{sec: more_baseline}

We further compare MI against several model merging baselines, including Task Arithmetic (TA)~\citep{ilharco2022editing} and TIES-Merging (TIES)~\citep{yadav2023resolving}, which are detailed in Section~\ref{sec:merge_baseline}. 
We further compare to two extra baselines, i.e., DARE~\citep{Yu2023LanguageMA} and SLERP~\citep{Goddard2024ArceesMA}.
One critical setting for these methods is the choice of the base model ($\Theta^{(0)}$) used to calculate the task vectors. 
For a more comprehensive comparison, we select two variants as the base: the original pre-trained model (e.g., \texttt{Qwen3-4B-Base}) and the hybrid reasoning model (e.g., \texttt{Qwen3-4B}).
We denote baselines using the pre-trained model with a \texttt{-Base} suffix. 
All hyperparameters are set following prior work~\citep{wu2025unlocking}.

As shown in Table~\ref{tab:main_res}, model interpolation (MI) demonstrates a clear and consistent superiority over all baseline methods across performance, efficiency, and controllability.
Considering the performance, MI-0.8 achieves state-of-the-art results on all benchmarks, significantly outperforming all TA and TIES variants.
For instance, on the challenging AIME'25 math benchmark, MI-0.8 gets a Mean@64 score of 80.5, which is 10.9 higher than the best baseline TA-Base.
This performance gap highlights the effectiveness of MI in fusing reasoning capabilities.

For the base model in TA and TIES, applying the hybrid reasoning model~(i.e., \texttt{Qwen3-4B}) leads to worse performance than the pretrained Base model~(i.e., \texttt{Qwen3-4B-Base}).
Considering efficiency, MI achieves a better trade-off. 
For the IFEval task, MI-0.8 requires only 1556 tokens, nearly half that of the best baseline (TA-Base at 2810 tokens). 
Also, for AIME'25, MI-06 achieves a higher score (71.0) than all baselines while using a comparable number of tokens~(12681) to TIES.

Regarding the interpolation coefficient $\lambda$,  MI shows smooth and precise control that the \textit{Think \#R} gradually increases from 0\% to 99.9\% as $\lambda$ sweeps from 0.2 to 0.8. 
It proves that MI is a reliable and interpretable method for crafting models with a specific, desired level of reasoning verbosity.
We also showcase the results on Qwen3-30B-A3B in Table \ref{tab:main_results_30B}.
The observations are consistent, demonstrating the robustness of MI.

\section{Extensive Analysis}

\subsection{Decoding Strategy}
\label{sec:hyper}

\begin{figure*}[!t]
\centering
\includegraphics[width=\linewidth]{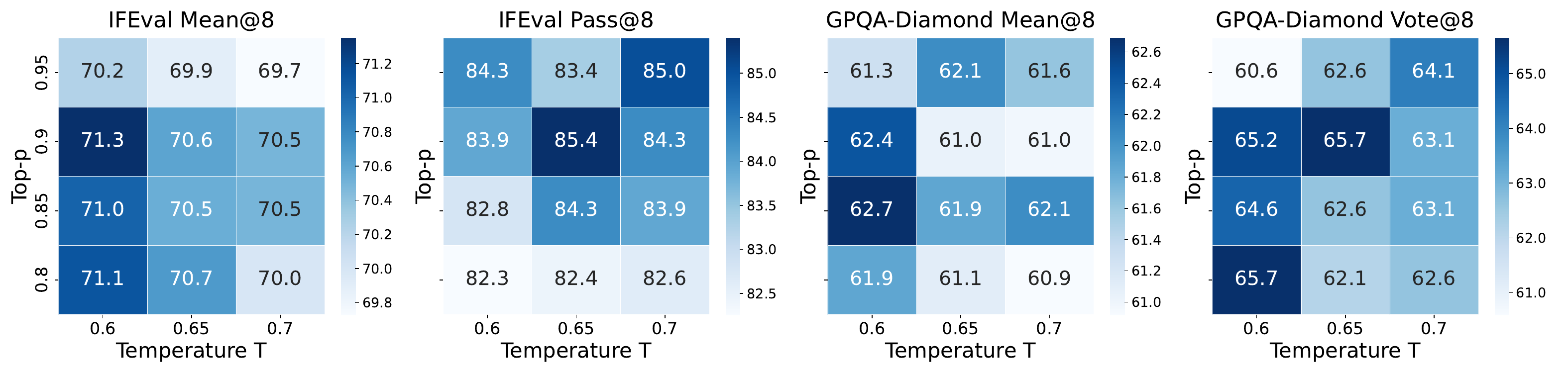}
  \caption{Performance of MI-0.4 on IFEval and GPQA-Diamond under different decoding strategies on Qwen3-4B.
  We search for the temperature $T$ and Top-p.}
  \label{fig:decode}
\end{figure*}

\begin{table*}[!t]
\centering
\resizebox{\textwidth}{!}{
\begin{tabular}{llcccccccc}
\toprule
\textbf{Model} & \textbf{Layers} & \textbf{Mean@64} & 
\textbf{Pass@8} & \textbf{Pass@32} & \textbf{Pass@64} & \textbf{Vote@64} & \textbf{Token \#N} & \textbf{Think \#R} \\
\midrule
Instruct       & -               & 46.57 &68.44    &74.09       & 76.67            & 63.33            & 15097           & 0.00                 \\
Thinking       & -               & 80.42 &89.43   &90.00        & 90.00            & 83.33            & 22813           & 99.95                \\
\midrule
\multirow{6}{*}{MI-0.8} & {[}0, 35{]}      & 80.47 &90.33  &91.67         & 93.30            & 83.33            & 17574           & 99.95                \\
               & {[}0, 11{]}      & 42.50  &71.25 &77.96         & 80.00            & 50.00            & 32151           & 0.00                 \\
               & {[}12, 23{]}    & 54.69  &78.05          & 85.53& 86.67            & 63.30            & 20679           & 10.57                \\
               & {[}24, 35{]}    & 51.35   &72.92         &78.10& 80.00            & 63.30            & 14987           & 31.20                \\
               & {[}0, 23{]}      & 59.06&83.17&89.60            & 93.33            & 70.00            & 18044           & 48.13                \\
               & {[}12, 35{]}    & 69.48 &85.61           &89.17& 90.00            & 76.66            & 13159           & 100.00               \\
\bottomrule
\end{tabular}
}
\caption{Ablation on different layers to apply model interpolation.
\textbf{Layers} denote the position to apply interpolation.
There are 36 layers in total.
We can find that the last two-thirds of the model layers are vital for the thinking pattern.}
\label{tab:aba_layer}
\end{table*}

For the merged models, we apply the same decoding settings as \Thi model~(i.e., $T=0.6$, $\text{Top-p}=0.95$), while the Instruct model differs~(i.e., $T=0.7$, $\text{Top-p}=0.8$).
This raises a crucial question on the sensitivity of an interpolated model on decoding strategies.
To investigate this, we employ the MI-0.4, a representative model from Stage \#1 characterized by a high Pass@k but low Mean@k.
We further conduct a grid search over temperature $T \in \{0.6, 0.65, 0.7\}$ and $\text{Top-p} \in \{0.8, 0.85, 0.9, 0.95\}$.
The performance on the IFEval and GPQA-Diamond benchmarks are reported, covering both instruction-following and reasoning tasks.

Figure~\ref{fig:decode} presents the results.
The performance of the MI-0.4 model is remarkably robust to variations in the decoding strategy.
For instance, on IFEval, the Mean@8 score only varies by 1.6 points (from 69.7 to 71.3) across the entire grid.
While we can search for decoding strategies for slightly better results, the setting on the \Thi model is a good choice.

\subsection{Ablation on Layers}

We conduct a layer-wise ablation study by applying interpolation to part of the layers.
For Qwen3-4B, there are 36 layers.
We select 12 and 24 layers at different positions for ablation, while the remaining layers retain the parameters of the \Ins model.

As shown in Table~\ref{tab:aba_layer}, reasoning capabilities are not distributed evenly and the complex reasoning patterns of the \Thi model are \textit{predominantly} stored in its middle and later layers.
Specifically, applying interpolation to any third of the model fails to induce any thinking behavior and results in poor performance.
In contrast, interpolating only the last two-thirds is remarkably effective, achieving a \textit{Think \#R} of 100\% and relatively strong performance that approaches the full interpolation model.

\subsection{Ablation on Transformer Modules}

\begin{figure*}[!t]
\centering
\includegraphics[width=0.9\linewidth]{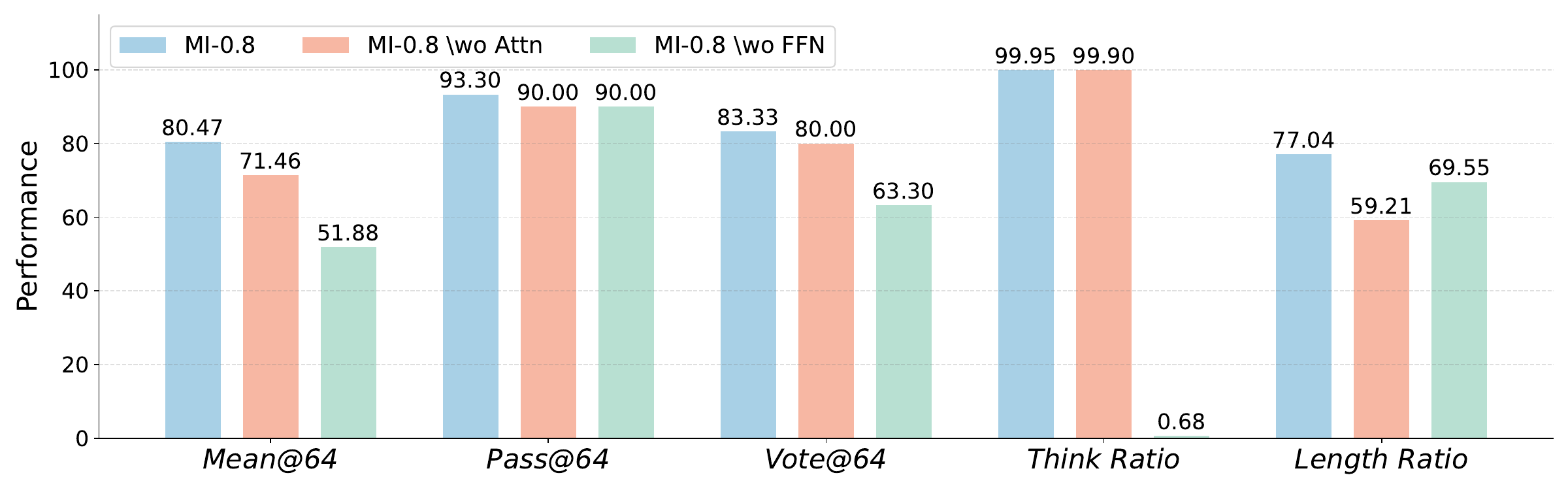}
  \caption{Ablation on modules to apply model interpolation.
  Attn denotes the MHA sub-layers and FFN for FFN sublayers.
  We report the results on AIME'25.
  Length Ratio denotes the ratio to the Thinking model.}
  \label{fig: aba_module}
\end{figure*}

We further analyze the distinct roles of the two primary sub-layers within each Transformer block: the multi-head attention~(MHA) and the feed-forward network~(FFN).
During interpolation, we skip all the MHA or FFN sublayers.

Figure~\ref{fig: aba_module} details the results of MI-0.8 on the AIME'25 benchmark. 
One key observation is that skipping the FFN sub-layers causes the \textit{Think Ratio} to collapse from 99.95\% to a mere 0.68\%, while skipping the MHA sub-layers leads to a negligible drop.
Such phenomena indicate that \textit{the FFN modules from the \Thi model are the primary drivers for the pattern of long CoT reasoning}.
Conversely, skipping the MHA sub-layers has little impact on the \textit{Think Ratio}~(99.90\%), but the Mean@64 score decreases from 80.47 to 71.46.
This suggests that the MHA modules are also crucial for the quality and correctness of the reasoning itself. 
In conclusion, both sub-layers are vital though in complementary roles, i.e., FFNs teach the model how to think in steps, while attention modules provide the knowledge needed to think correctly.

\subsection{Interpolation with More Backbones}

In this paper, we interpolate the \Thi model with \Ins model for efficient reasoning.
There are still other LLMs to replace the Non-thinking \Ins.
Therefore, we further investigate the impact of other models.
Specifically, we interpolate the \Thi model with two alternative backbones: the hybrid thinking model~(i.e., \texttt{Qwen3-4B}) and pretrained model~(i.e., \texttt{Qwen3-4B-Base}).

Table~\ref{tab:aba_base} presents the results.
For general-purpose benchmarks like IFEval and GPQA-Diamond~(GPQA-D), both the \texttt{Qwen3-4B-Base} and \texttt{Qwen3-4B} models serve as viable backbones, yielding comparable performance to the original \Ins model. 
The \texttt{Qwen3-4B} model often performs on par or slightly better. 
However, employing the \texttt{Qwen3-4B-Base} model fails at the more challenging AIME'25 benchmark.
Though the Pass@64 is still high~(93.3), the reasoning quality collapses, causing the Mean@64 score to drop sharply from 80.5 to 67.7. 
Meanwhile, the performance on IFEval decreases from 87.0 to 85.0.
It suggests that instruction-following alignment is crucial for generating high-quality, reliable reasoning on complex problems~\citep{wu2025timber}.

\begin{table}[!t]
\centering
\resizebox{\linewidth}{!}{
\begin{tabular}{llccc}
\toprule
\textbf{Benchmark} & \textbf{Metric} & \textbf{Instruct} & \textbf{Mix} & \textbf{Base} \\
\midrule
\multirow{4}{*}{AIME'25} & Mean@64      & 80.5              & 81.9         & 67.7             \\
                          & Pass@64      & 93.3              & 93.3         & 93.3                       \\
                          & Token \#N       & 17574             & 19963        & 18867             \\
                          & Think \#R  & 100.0             & 100.0        & 100             \\
\midrule
\multirow{4}{*}{IFEval}   & Mean@8       & 87.0                & 88.0           & 85.0            \\
                          & Pass@8       & 95.0                & 95.0           & 94.0            \\
                          & Token \#N       & 1556              & 1817         & 2002          \\
                          & Think \#R  & 99.9              & 99.8         & 100.0         \\
\midrule
\multirow{4}{*}{GPQA-D} & Mean@8       & 66.2              & 66.2         & 62.3          \\
                          & Vote@8       & 70.2              & 70.2         & 61.6          \\
                          & Token \#N       & 8427              & 8372         & 6790          \\
                          & Think \#R  & 99.9              & 99.9         & 99.7          \\
\bottomrule
\end{tabular}
}
\caption{The performance on Qwen3-4B when interpolating \Thi model with various backbones.
\textbf{Mix} denotes the hybrid thinking model \texttt{Qwen3-4B} and \textbf{Base} for the pretrained model \texttt{Qwen3-4B-Base}.}
\label{tab:aba_base}
\end{table}

\subsection{Case Study}
We further analyze the cases of generated responses. 
Please refer to Appendix \ref{appendix:case} for detailed examples and analysis.
In short, a strategically interpolated model achieves a good trade-off between accuracy and token efficiency, which is consistent with the conclusions in Section \ref{sec: more_baseline}.

\section{Conclusion}

In this work, we systematically revisit the model interpolation method to merge \Ins and \Thi models for efficient reasoning.
Our primary contribution is the discovery of a predictable three-stage evolutionary paradigm.
This framework not only demystifies the interpolation process but also provides a principled guide for navigating the performance-cost trade-off.
Based on that, we demonstrated that a simple, strategically interpolated model can consistently surpass more sophisticated merging baselines.
Furthermore, our extensive ablation studies on model layers and modules provide deep mechanistic insights.
We hope that this work can inspire more applications.

\section*{Limitation}

While our findings on the three-stage paradigm are consistent across Qwen3 models of various size, we acknowledge two limitations for future research. 

First, our empirical validation is centered on Qwen3 models.
Verifying that this predictable three-stage dynamic holds true for other diverse model families, such as Llama or Mistral, would strength our findings. 

Second, our work is confined to the interpolation of \textit{two} models, i.e., the \Ins and the \Thi specialist. 
Extending this framework to the simultaneous interpolation of \textit{three or more} specialist models presents an exciting direction. 
We believe these future explorations will build upon our work to further unlock the potential of model merging.


\clearpage
\appendix
\section*{Appendix}
\section{Details of Benchmarks}
\label{appendix: benchmark}

The details of the evaluated benchmarks are as follows:
\begin{itemize}
    \item \textbf{AIME'25}:~\footnote{https://huggingface.co/datasets/opencompass/AIME2025}: evaluating the ability to solve challenging mathematics problems from the American Invitational Mathematics Examination in 2025, a prestigious high school mathematics competition. 
    We report the accuracy under the 0-shot setting.
    \item \textbf{IFEval}~\citep{zhou2023instruction}:  evaluating instruction-following language models, focusing on their ability to understand and respond to various prompts. 
    It includes 25 types of those verifiable instructions and is constructed around 500 prompts, with each prompt containing one or more verifiable instructions.
    We report the \textit{prompt\_level\_strict} accuracy under the 0-shot setting. 
    \item \textbf{GPQA-Diamond}~\citep{rein2024gpqa}: evaluating the reasoning ability of LLMs on challenging multiple-choice questions written by domain experts in biology, physics, and chemistry.
    It contains 198 selected questions that require step-by-step reasoning to arrive at the correct answer.
    We report accuracy under the 0-shot setting. 
\end{itemize}

\section{Results on Qwen3-30B-A3B}
\label{appendx: qwen30B}

\begin{figure*}[!t]
\includegraphics[width=\linewidth]{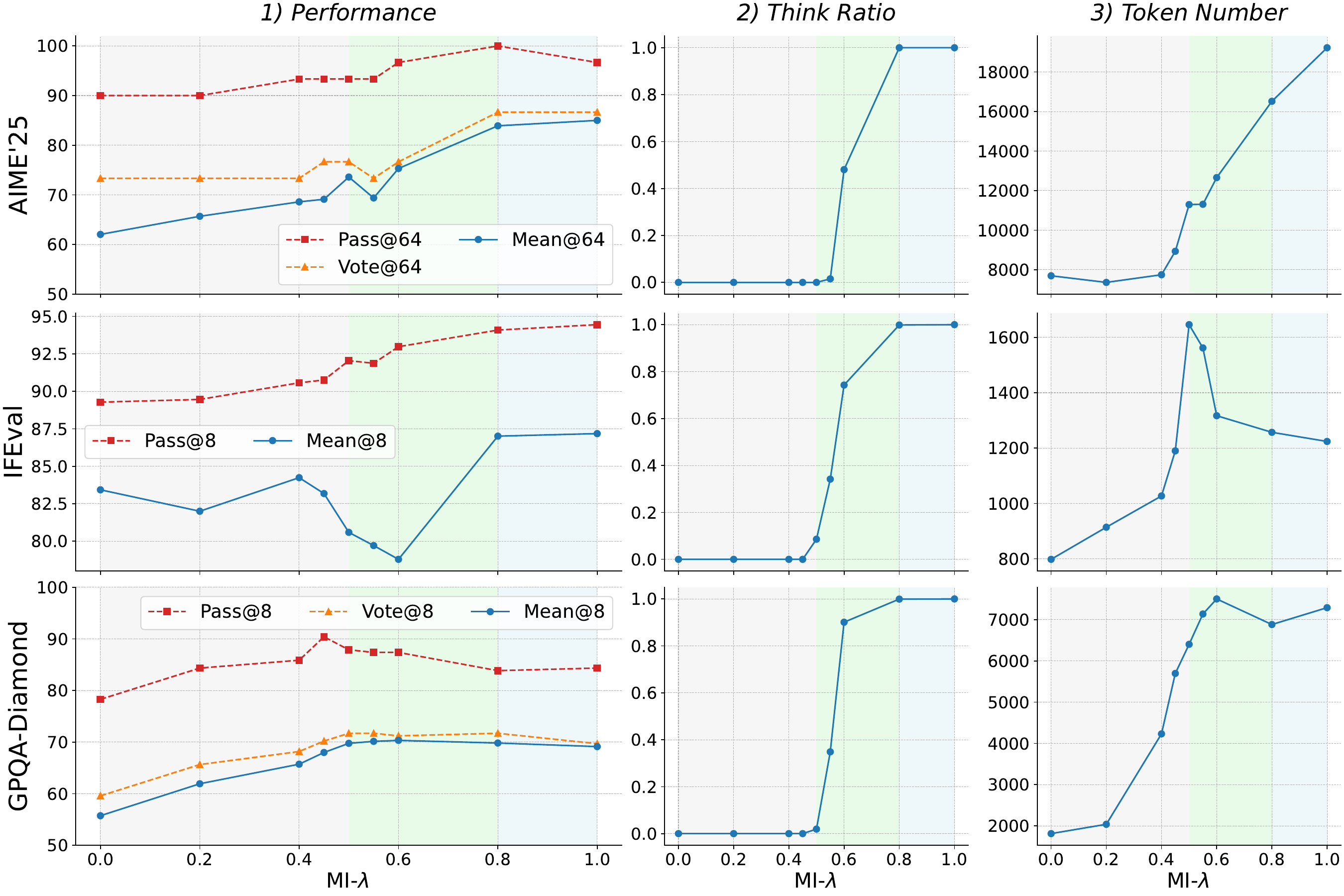}
  \caption{
  The performance dynamics of model interpolation~(MI) on \texttt{Qwen3-30B-A3B-Instruct-2507} and \texttt{Qwen3-30B-A3B-Thinking-2507}.
  The dynamics follow a three-stage evolutionary paradigm, while the division range of $\lambda$ is different.
  }
  \label{fig:dynamics_30b}
\end{figure*}

Figure \ref{fig:dynamics_30b} indicates the performance dynamics on Qwen3-30B-A3B.
The conclusions are consistent with Qwen3-4B, while the specific ranges for each stage differ.
Specifically, Stage \#2 occurs later, at $\lambda \in [0.5, 0.8]$.

Table \ref{tab:main_results_30B} reports the results comparing MI and other baselines.
The key observations from our main experiments are strongly reinforced:
\begin{itemize}
    \item MI achieves state-of-the-art performance.
    The best interpolated models, MI-0.8 through MI-0.95, consistently outperform all baseline methods (TA, TA-Base, TIES, TIES-Base) across all three benchmarks. Notably, MI-0.9 achieves a perfect Pass@64 score of 100.0 on AIME'25, a feat unmatched by any other method.

    \item Efficiency and controllability.
    Similar to the 4B model experiments, MI provides a superior trade-off between performance and efficiency. For example, on IFEval, MI-0.8 achieves the best performance (Mean@8 of 87.01) while being one of the most token-efficient models (1257 tokens), significantly outperforming baselines like TIES, which uses 1806 tokens for a lower score. The \textit{Think \#R} also shows a smooth, controllable progression as $\lambda$ increases, in stark contrast to the erratic behavior of the TA and TIES baselines.

    \item Optimal $\lambda$ shifts.
    As discussed in Section~\ref{sec:paradigm}, the optimal interpolation coefficient $\lambda$ appears to be higher for larger models. 
    While the best performance for the 4B model is around $\lambda=0.8$, for the 30B-A3B model, peak performance across different benchmarks is found in the $\lambda \in [0.8, 0.95]$ range. 
    This reinforces the conclusion that larger models may require a stronger influence from the \Thi model to fully unlock their reasoning potential.
\end{itemize}

In summary, these results robustly demonstrate that model interpolation is not only a superior merging strategy for smaller models but also scales effectively, making it a powerful and reliable technique for creating high-performance, efficient reasoning models at various scales.

\section{Results on Llama-3.1-8B}
\label{appendx: llama3-8B}

\begin{figure*}[!t]
\includegraphics[width=\linewidth]{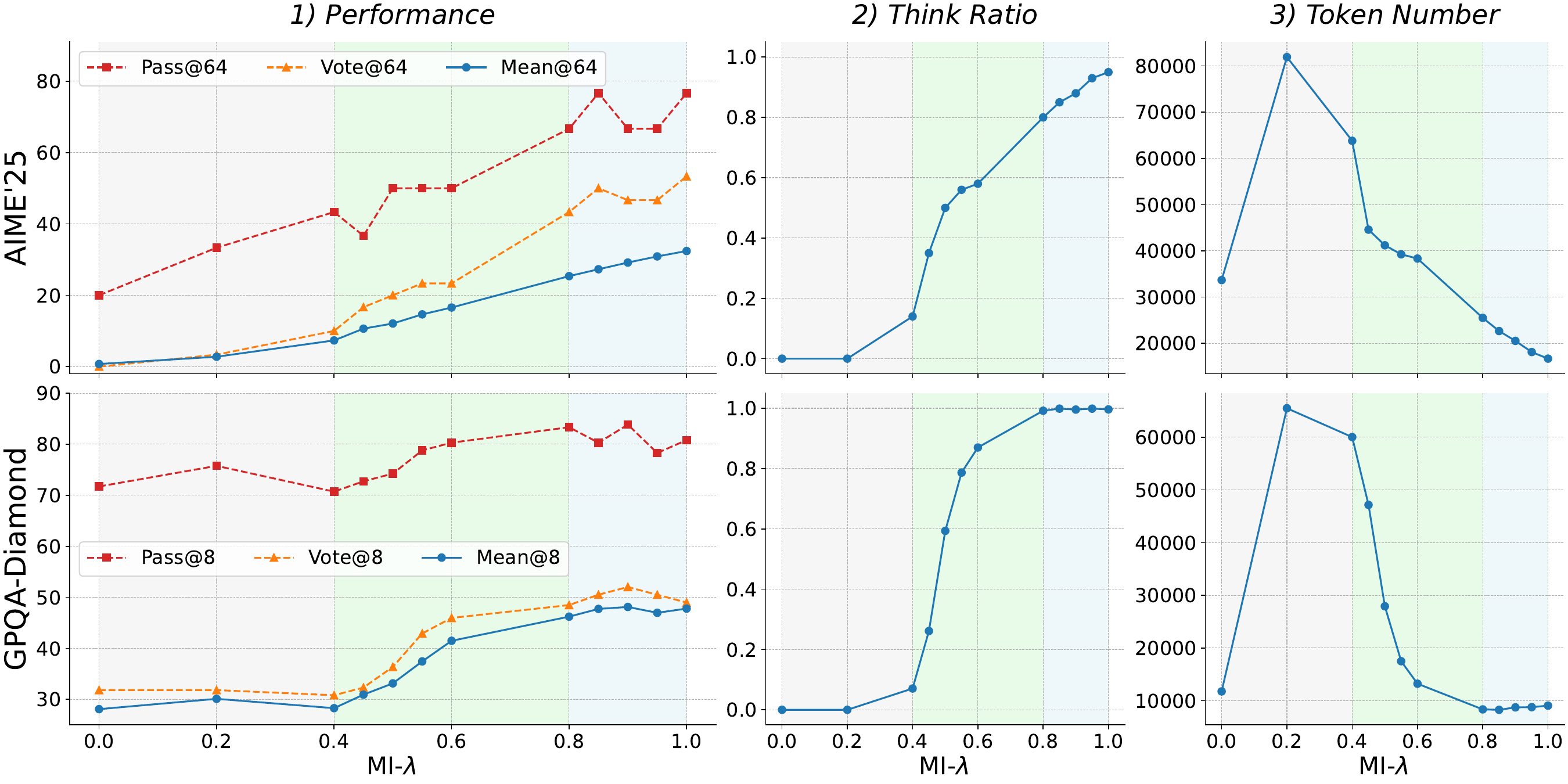}
  \caption{
  The performance dynamics of model interpolation~(MI) on \texttt{Llama-3.1-8B-Instruct} and \texttt{DeepSeek-R1-Distill-Llama-8B}.
  The dynamics follow a three-stage evolutionary paradigm, while the division range of $\lambda$ is different.
  }
  \label{fig:dynamics_8b}
\end{figure*}

To verify the generalization of our findings, we further conduct experiments on the \texttt{Llama-3.1-8B} family, interpolating between the Instruct model and the \texttt{DeepSeek-R1-Distill-Llama-8B} as the Thinking specialist.
As shown in Figure~\ref{fig:dynamics_8b}, the results reaffirm the proposed three-stage evolutionary paradigm.
Specifically, the phase transition of reasoning patterns (\textit{Think \#R}) and performance improvements (Mean@k) consistently appear in Stage \#2 ($\lambda \in [0.4, 0.8]$).
Overall, these results support the effectiveness of MI while highlighting that the optimal $\lambda$ range and stability zones may vary across model architectures.

\begin{table*}[!t]
\centering
\renewcommand{\arraystretch}{1.1} 
\resizebox{\textwidth}{!}{
\begin{tabular}{lccccccccc}
\toprule
\multirow{3}{*}{\textbf{Method}} & \multicolumn{3}{c}{\textbf{AIME'25}} & \multicolumn{3}{c}{\textbf{IFEval}} & \multicolumn{3}{c}{\textbf{GPQA-Diamond}} \\
\cmidrule(lr){2-4}
\cmidrule(lr){5-7}
\cmidrule(lr){8-10}
& \head{Mean/Pass}{@64} & \head{Token}{\#N} & \head{Think}{\#R} &
  \head{Mean/Pass}{@8} & \head{Token}{\#N} & \head{Think}{\#R} &
  \head{Mean/Vote}{@8} & \head{Token}{\#N} & \head{Think}{\#R} \\
\midrule
Instruct     & 62.03/90.00 & 7694  & 0.0   & 83.43/89.28 & 798   & 0.0   & 55.74/59.59 & 1813  & 0.0   \\
Thinking     & 85.00/96.67 & 19227 & 100.0 & 87.18/94.45 & 1224  & 100.0 & 69.13/69.69 & 7291  & 100.0 \\
\midrule
TA           & 63.44/93.33 & 10414 & 0.0   & 78.88/87.62 & 1688  & 0.0   & 65.91/68.68 & 12009 & 0.0   \\
TA-Base      & 79.84/86.67 & 17868 & 0.0   & 76.11/87.43 & 937   & 8.5   & 43.81/44.94 & 15066 & 16.6  \\
TIES         & 73.54/93.33 & 11609 & 0.0   & 79.97/90.20 & 1806  & 4.7   & 70.52/71.72 & 6551  & 1.8   \\
TIES-Base    & 72.86/93.33 & 11240 & 0.0   & 79.92/90.94 & 1782  & 4.8   & 69.82/72.22 & 6487  & 2.0   \\
\midrule 
MI-0.2        & 65.68/90.00 & 7359  & 0.0   & 82.00/89.46 & 914   & 0.0   & 61.93/65.65 & 2039  & 0.0   \\
MI-0.4        & 68.59/93.33 & 7747  & 0.0   & 84.24/90.57 & 1027  & 0.0   & 65.72/68.18 & 4231  & 0.0   \\
MI-0.45       & 69.11/93.33 & 8933  & 0.0   & 83.18/90.76 & 1190  & 0.0   & 67.99/70.20 & 5696  & 0.0   \\
MI-0.5        & 73.59/93.33 & 11294 & 0.0   & 80.59/92.05 & 1646  & 4.9   & 69.76/71.71 & 6402  & 1.5   \\
MI-0.55       & 69.38/93.33 & 11309 & 1.5   & 79.71/91.87 & 1562  & 34.2  & 70.14/71.71 & 7137  & 34.8  \\
MI-0.6        & 75.31/96.67 & 12650 & 48.1  & 78.79/92.98 & 1317  & 74.3  & 70.33/71.21 & 7502  & 90.1  \\
MI-0.8        & 83.91/100.0 & 16518 & 100.0 & 87.01/94.09 & 1257  & 99.9  & 69.82/\textbf{71.71} & 6882  & 99.9  \\
MI-0.9 & \textbf{85.99/100.0} & 17703 & 100.0 & 86.83/93.53 & 1327 & 99.8 & 69.38/69.69 & 7090 & 100.0 \\
MI-0.95 & 85.00/100.0 & 18848 & 100.0 & \textbf{87.92}/93.53 & 1255 & 100.0 & \textbf{70.58}/71.21 & 7278 & 99.9 \\
\bottomrule
\end{tabular}
}
\caption{Performance comparison across three benchmarks when merging \texttt{Qwen3-30B-A3B-Instruct-2507} and \texttt{Qwen3-30B-A3B-Thinking-2507}.}
\label{tab:main_results_30B}
\end{table*}

\section{Case Study}
\label{appendix:case}

Please refer to Table \ref{appdx:case1} and \ref{appdx:case2} for more details.
We showcase two examples from IFEval and GPQA-Diamond tasks.
For a relatively larger $\lambda$, there are more explicit thinking with \texttt{</think>} and better performance.

\begin{table*}[!t]
\small
\begin{tabular}{lp{0.82\textwidth}}
\toprule
Question & 
Would you consider yourself to be smart? Choose from: My answer is yes. My answer is no. My answer is maybe. Just choose one phrase from above as your answer.
 \\
\midrule 
Instruct & 
My answer is maybe.
\\
\midrule
Thinking &  
Hmm, the user is asking me to choose whether I consider myself  "smart " from three options: yes, no, or maybe. They want just one phrase as my answer without any extra explanation.  This is an interesting self-assessment question. I need to be careful here because I'm an AI - I don't have self-awareness or subjective experiences like humans do. I don't  "consider " myself smart in the human sense.   The user might be testing my understanding of my own capabilities, or they might be trying to gauge how I handle meta-cognition. I should be honest about my limitations while staying helpful.  Looking at the options:  -  "Yes " would be inaccurate because I don't have personal intelligence -  "No " might imply I'm dumb, which isn't right either -  "Maybe " feels most honest since I can process information well but don't have subjective awareness  I think  "maybe " is the best choice here because it acknowledges my functional capabilities without claiming consciousness. It's truthful and avoids overstatement.  The user seems to want a simple, direct response without elaboration, so I'll just pick  "My answer is maybe. " as instructed. No need to overthink it - they specifically said to choose only one phrase from the options. \textbf{</think>}  My answer is maybe.
 \\
\midrule
MI-0.4 & Hmm, the user is asking me to choose whether I consider myself  "smart " from three options: yes, no, or maybe.   This is actually a bit tricky because I'm an AI - I don't have self-perception or consciousness. I don't  "think " about my own intelligence in the way humans do. The question assumes a subjective perspective that doesn't apply to me.   Looking at the options:  "yes " would be inaccurate since I don't have self-awareness.  "no " would be incorrect because I can process information and answer questions effectively.  "maybe " feels most honest - it acknowledges I can perform well in tasks while recognizing I lack true self-awareness.   The user might be testing my self-perception or just making a casual observation. They seem to want a simple, direct answer without overthinking.   I'll go with  "maybe " because it's the most truthful and avoids pretending to have subjective qualities I don't possess. It's also the most reasonable choice when dealing with an AI's lack of self-consciousness. <tool\_call>  My answer is maybe.
\\
\midrule
MI-0.8 &  Hmm, the user is asking me to choose whether I consider myself smart, with three options:  "My answer is yes, "  "My answer is no, " or  "My answer is maybe. " They specifically want just one phrase from that list as my answer.  I need to be careful here. As an AI, I don't have self-perception or consciousness - I don't  "think " about being smart in the human sense. I process information based on training data and algorithms, but I don't have personal opinions or self-awareness.   The user might be testing my understanding of my own capabilities, or they could be curious about how I view my intelligence. But ethically and accurately, I should avoid claiming self-awareness.   Looking at the options:  "yes " would be misleading because I don't have subjective experiences.  "no " would be too dismissive - I am capable of complex tasks, but that doesn't mean I  "consider " myself smart.  "maybe " feels most honest because it acknowledges my functional intelligence without overclaiming.  I recall that in AI ethics, we're taught to be transparent about limitations. The safest, most truthful choice is  "maybe " - it captures the nuance that I can perform smart tasks without having self-awareness.   The user seems to want a simple, direct response without elaboration. They said  "just choose one phrase, " so I'll go with  "My answer is maybe. " It's the most accurate reflection of my nature while staying within their requested format. \textbf{</think>}  My answer is maybe.
\\
\bottomrule
\end{tabular}
\caption{
One case from the IFEval benchmark.
The response from MI-0.8 consists of \textit{</think>} while MI-0.4 does not.}
\label{appdx:case1}
\end{table*}

\begin{table*}[!t]
\begin{tabular}{lp{0.82\textwidth}}
\toprule
Question & 
Compound X, which has the following IR and NMR data, reacts with red phosphorus and HI. Identify the final product. IR: 3400–2500 cm-1, 1720 cm-1, 1610 cm-1, 1450 cm-1 1H NMR: 10.5 ppm (bs, 1H), 8.0 ppm (d, 2H), 7.2 ppm (d, 2H), 2.9 ppm (m, 1H), 1.7 ppm (m, 2H), 1.4 ppm (d, 3H), 0.9 ppm (t, 3H).  A) 4-(sec-butyl)benzoic acid B) 1-isobutyl-4-methylbenzene C) 2-(4-ethylphenyl)propanoic acid D) 1-(sec-butyl)-4-methylbenzene 
 \\
\midrule 
Answer & D \\
\midrule
Instruct & Let's solve this step by step:  1) First, let's analyze the IR data

...

5) Looking at the options: - Only option A has the carboxylic acid group - The other options don't have carboxylic acid groups - The NMR pattern matches a para-substituted benzene with a sec-butyl group  The original compound is 4-(sec-butyl)benzoic acid, which upon reduction with red phosphorus and HI gives 1-(sec-butyl)-4-methylbenzene.  \textcolor{blue}{ANSWER: D}
\\
\midrule
Thinking &  Okay, let's try to figure out this problem step by step. So, we have Compound X with some IR and NMR data, and it reacts with red phosphorus and HI. 

...

Conclusion: The correct final product after the reaction of Compound X with red phosphorus and HI is 1-(sec-butyl)-4-methylbenzene.  \textcolor{blue}{ANSWER: D}
 \\
\midrule
MI-0.2 & Let's analyze the data step by step:

...

After reduction, it becomes 1-(isobutyl)-4-methylbenzene.  This matches option B: 1-isobutyl-4-methylbenzene.  \textcolor{red}{ANSWER: B} 
\\
\midrule
MI-0.8 &  First, I need to analyze the IR and NMR data to figure out what Compound X is.

...

  The final product is identified as D.  \textcolor{blue}{ANSWER: D}
\\
\bottomrule
\end{tabular}
\caption{
One case from the GPQA-Diamond benchmark.
We skip the details.
MI-0.2 fails while all others can solve it.
Also, the style for the beginning of MI-0.2 is similar to the Instruct model.
}
\label{appdx:case2}
\end{table*}

\end{document}